\documentclass[letterpaper, 10 pt, conference]{ieeeconf}  % Comment this line out if you need a4paper

\IEEEoverridecommandlockouts                              % This command is only needed if 
\usepackage{graphics} % for pdf, bitmapped graphics files
\usepackage{epsfig} % for postscript graphics files
\usepackage{times} % assumes new font selection scheme installed
\usepackage{amsmath} % assumes amsmath package installed
\usepackage{amssymb}  % assumes amsmath package installed
\usepackage{arydshln}
\usepackage{multirow}

\usepackage{xcolor}
\usepackage{soul}
\usepackage{fixltx2e}

\usepackage[nobiblatex]{xurl}

\usepackage{color,soul}

\title{\LARGE \bf
Reducing Annotating Load: Active Learning with Synthetic Images in Surgical Instrument Segmentation
}

\author{Haonan Peng, Shan Lin, Daniel King, Yun-Hsuan Su, Randall A. Bly, Kris S. Moe, Blake Hannaford % <-this % stops a space
\thanks{Haonan Peng, Shan Lin, Daniel King, Randall A. Bly, Kris S. Moe and Blake Hannaford are with University of Washington, Seattle, WA 98195, USA.
        {\tt\small {penghn}@uw.edu}}%
\thanks{Yun-Hsuan Su is with Mount Holyoke College, South Hadley, MA 01075, USA.
        {\tt\small msu@mtholyoke.edu}}%
}

\begin{document}
\bstctlcite{IEEEexample:BSTcontrol}

\maketitle
\thispagestyle{empty}
\pagestyle{empty}

%%%%%%%%%%%%%%%%%%%%%%%%%%%%%%%%%%%%%%%%%%%%%%%%%%%%%%%%%%%%%%%%%%%%%%%%%%%%%%%%
\begin{abstract}

Accurate instrument segmentation in endoscopic vision of robot-assisted surgery is challenging due to reflection on the instruments and frequent contacts with tissue. Deep neural networks (DNN) show competitive performance and are in favor in recent years. However, DNN’s hunger for labeled data poses a huge workload of annotation. Motivated by alleviating this workload, we propose a general embeddable method to decrease the usage of labeled real images, using active generated synthetic images. In each active learning iteration, the most informative unlabeled images are first queried by active learning and then labeled. Next, synthetic images are generated based on these selected images. The instruments and backgrounds are cropped out and randomly combined with each other with blending and fusion near the boundary. The effectiveness of the proposed method is validated on 2 sinus surgery datasets and 1 intraabdominal surgery dataset. The results indicate a considerable improvement in performance, especially when the budget for annotation is small. The effectiveness of different types of synthetic images,  blending methods, and external background are also studied. All the code is open-sourced at: https://github.com/HaonanPeng/active\_syn\_generator. 
\end{abstract}

%%%%%%%%%%%%%%%%%%%%%%%%%%%%%%%%%%%%%%%%%%%%%%%%%%%%%%%%%%%%%%%%%%%%%%%%%%%%%%%%
\section{Introduction}

Minimally invasive surgery (MIS) has seen rapid development in recent years in applications such as intra-abdominal surgery and otolaryngology, and can improve the patients’ outcome and recovery \cite{sayari2019review}, \cite{peters2018review}. In MIS, endoscopes are used to provide vision of the surgical site in real-time. One of the most important components of understanding endoscopic surgical images is the segmentation of instruments, and much recent research is applying deep learning technology \cite{maier2017surgical}, \cite{shvets2018automatic}. However, the lack and cost of labeled data still remains a major challenge for many learning-based methods, especially in medical practice where the resource of data is limited and sometimes only trained experts can annotate images with high quality \cite{cheplygina2019not}, \cite{yang2017suggestive}.

Recently, using synthetic data to alleviate the workload of annotating data draws more attention, especially image-to-image translation techniques such as generative adversarial networks (GANs) \cite{wang2021review}. Synthetic images generated from simulation have accurate labels without manual work, but the domain gap between synthetic and real datasets can reduce the accuracy of a model trained by synthetic images when evaluated on real images. Thus, domain randomization or domain adaptation is usually needed to address this gap \cite{tobin2017domain}, \cite{kouw2019review}. Although their performance is competitive, properly setting up GAN-based simulation environments for each application still remains a considerable amount of work.  

In contrast, the method of generating synthetic images by cutting and pasting generalizes to many segmentation tasks. For endoscopic sinus surgery, the reflections on metallic instruments, as well as blur, liquids, and occlusion on the tissue-instrument boundary makes it even harder to perform image-to-image translation \cite{lin2020lc}. An alternative to  generating synthetic images from simulation is copying and pasting real object images onto real background images. This is proven to be an efficient method to generate synthetic images without concern of domain gap \cite{dwibedi2017cut}. Compared with simulation, this method requires a certain amount of labeled data.

Another popular method to reduce the usage of real labeled data is active learning(AL) \cite{budd2021survey}, when combined with deep learning, it  achieves faster convergence and increased performance with fewer data, and goes back as far as 1988 \cite{angluin1988queries}. AL uses query criterion to select the most uncertain or informative samples from an unlabeled data set and ask for annotating \cite{gorriz2017cost}. This is suitable for surgical instrument segmentation because there are plenty of unlabeled videos but annotation is expensive \cite{qin2020towards}. 

In this work, we develop a copy-and-paste method to generate synthetic images, combined with active learning. We use active learning to choose the most informative unlabeled images to annotate, and use the copy-and-paste method to generate synthetic images which can ‘make best use of’ the selected real images. Then we experimentally show that the segmentation model trained with the synthetic images and a smaller number of selected real images have competitive performance compared to those trained on fully labeled real datasets. Three open source datasets are used in the experiments - UW-Sinus-Surgery-C/L Dataset \cite{qin2020towards} and EndoVis 2017 Dataset \cite{allan20192017}.

%%%%%%%%%%%%%%%%%%%%%%%%%%%%%%%%%%%%%%%%%%%%%%%%%%%
\section{Related Work}
\begin{figure*}
\centering
\vspace{0.3em}
\includegraphics[width=0.9\textwidth]{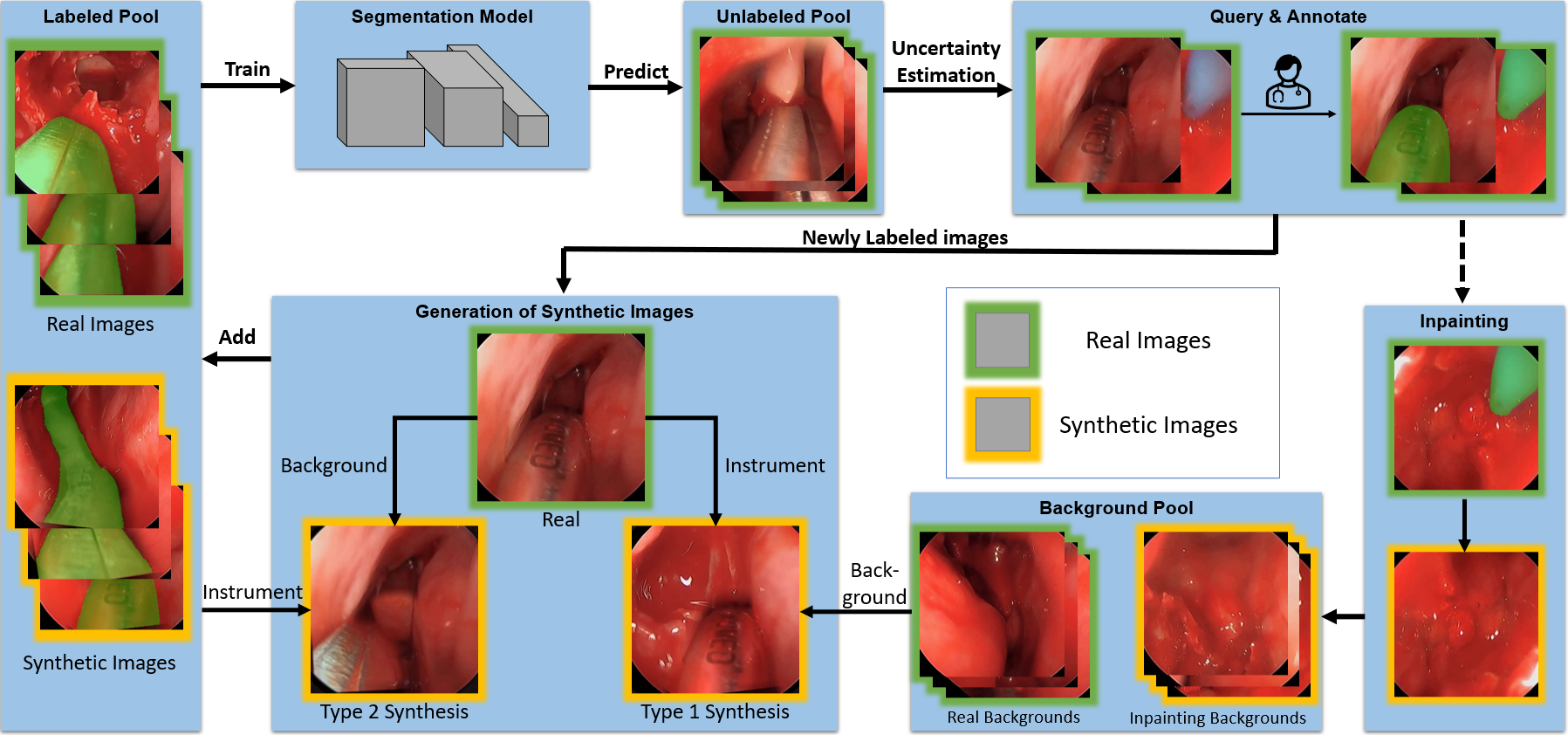}
\vspace{-1.em}
\caption{Workflow of the system} 
\vspace{-1.5em}
\label{workflow}
\end{figure*}
With the vigorous development of deep learning in recent years, cutting-edge performance in surgical instrument segmentation is achieved by deep convolutional neural networks \cite{qin2020towards}, \cite{islam2019learning}, \cite{kalinin2020medical}. However the Deep models’ hunger for large amounts of labeled data draws attention to reducing the workload of annotation, especially in medical image segmentation \cite{rajotte2021reducing}, \cite{fujita2020deep}. Using synthetic images is an intuitive approach. Among  approaches for generation of accurate, labeled synthetic images, copying the target object from one image and pasting it onto another image is feasible and relatively easy to implement. Dwibedi et al. propose a ‘cut’ and ‘paste’ method to generate synthetic images and train neural networks for kitchen object detection. Their result shows that just simply copying and pasting can result in artifacts such as aliasing of boundaries which decreases the learning performance. By improving the blending, synthetic data combined with ${10\%}$ real data can reach a competitive performance \cite{dwibedi2017cut}. Ghiasi et al. proposed a study on copy-paste data argumentation for instance segmentation which described a simple copy-pasting mechanism that improved the performance of strong baselines and could also be combined with semi-supervised methods such as self-training. Unlike \cite{dwibedi2017cut}, the study shows that simple pasting without any blending has similar performance with blended ones \cite{ghiasi2021simple}. Remez et al. present an object instance segmentation with a weakly-supervised cut-and-paste adversarial learning. Detection boxes and Faster R-CNN features are passed into a mask generator, which outputs segmentation masks that can be used to cut-and-paste the object into a new image location. And then a discriminator tries to distinguish between real and synthetic images, and select for training the ones that can improve learning performance \cite{remez2018learning}. GANs \cite{goodfellow2020generative} are also implemented to generate synthetic medical images\cite{singh2021medical} \cite{yoo2020generative}. GANs train two networks simultaneously - a generator and a discriminator, where the generator is trained to generate synthetic images that can cheat the discriminator, meanwhile the discriminator is trained to distinguish the real images and synthetic images. Recent implementations include image-to-image translation from simulated images to real images\cite{colleoni2021robotic}, and from cadaver images to live images\cite{lin2020lc}.  

Active learning is dedicated to selecting and labeling the most informative training images that can reach near-optimal performance with the fewest annotations (human effort) \cite{tajbakhsh2020embracing}, \cite{kim2020active}. Typically, in active learning, unlabeled images are selected by criteria such as maximum entropy or least confidence \cite{holub2008entropy}, \cite{roels2019cost}, \cite{schein2007active}. In some cases, however, these criteria do not outperform random selection \cite{yang2017suggestive}, \cite{belharbi2021deep}. Thus, more advanced criteria such as Bayesian active learning by disagreement (BALD) are proposed \cite{houlsby2011bayesian}. In BALD, Bayesian networks can be obtained by applying Monte Carlo dropout (MC dropout) to the network to generate a population of modified networks.  The BALD criterion combines a high overall uncertainty with a term which increases weight on disagreement among the population. Saidu et al. presented a study on semantic segmentation of prostate medical images with active learning, and the BALD criterion outperformed maximum entropy, especially when the budget for annotation is small \cite{gal2017deep}. Tran et al. proposed a Bayesian generative
active deep learning, which combines active learning and GAN data augmentation. The evaluation on image classification tasks suggests that the combined method outperform each single method.

In our proposed method, copy-and-paste synthetic images are generated based on informative real images selected by AL. Segmentation models trained by these images outperforms the models trained by only AL chosen real images or randomly generated synthetic images. In synthetic images, fusion near the boundary of the instrument reduces the artifact caused by copy-and-paste and thus improves the performance of segmentation near the boundary. By modifying several parameters, the proposed method is easily generalized to three different datasets.

%%%%%%%%%%%%%%%%%%%%%%%%%%%%%%%%%%%%%%%%%%%%%%%%%%%%%%%%%%%%%%%%%%%
\section{Methods}

\begin{figure*}
\centering
\vspace{0.3em}
\includegraphics[width=0.7\textwidth]{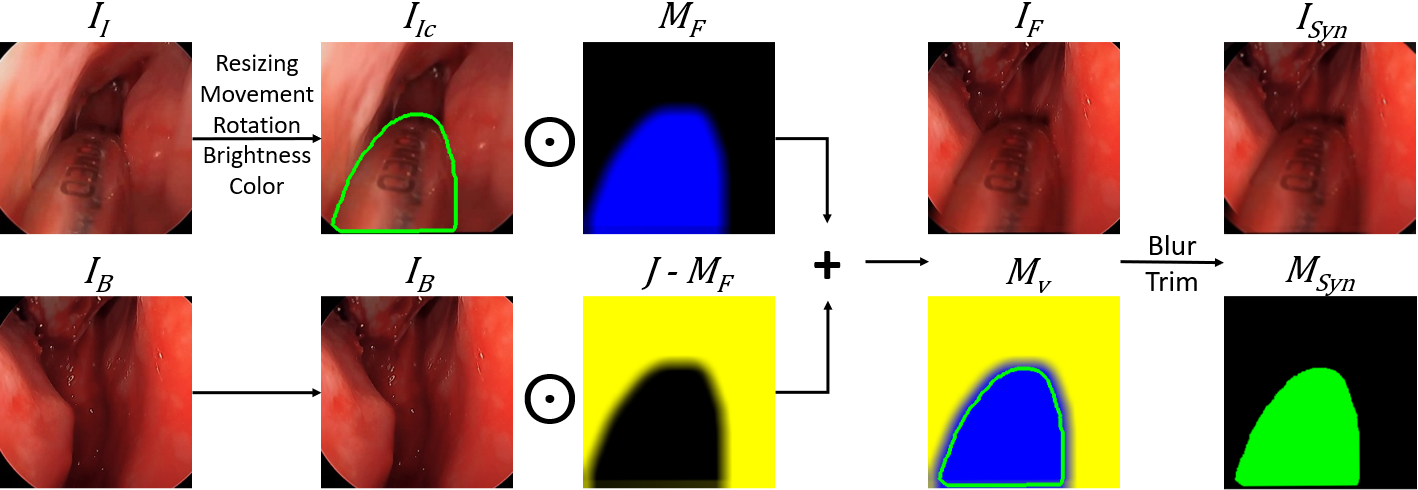}
\vspace{-1.em}
\caption{Generation of synthetic image. Please notice that $M_v$ is only for visualization, where the solid green line indicates the outline of the instrument,  yellow area comes from the background image $I_B$ and blue area comes from the instrument image $I_{IC}$. Transition area can be found around the boundary of the instrument on the synthetic image.} 
\vspace{-1.5em}
\label{gen_syn}
\end{figure*}

\subsection{System Workflow} \label{sch_sys_workflow}

Fig. \ref{workflow} shows the workflow of the entire system (details below). The overall objective of the proposed system is to use active learning to choose the most informative samples from the unlabeled pool of the database and ask for annotation. And then, synthetic images are generated and added to the labeled pool together with the real images to ‘make best use of’ the real images which are identified as most informative. The goal is to train a segmentation model on fewer labeled real images but having competitive performance.

The database (described in detail in \ref{sch_datasets}) consists of a labeled pool and an unlabeled pool of images. In terms of semantic segmentation of videos of endoscopic robotic surgery, it is not difficult to obtain unlabeled videos and images. Some of the datasets contain background images in which surgical tools are not presented. Compared to labeling the mask of instruments, it is not costly to manually select the background-only images to form the background pool.

Initially, some real images are randomly chosen and moved from the unlabeled pool to the labeled pool with annotation.  Synthetic images are first generated using the images in the labeled pool and are then added to the labeled pool. Next, a segmentation model is trained using the labeled pool and then makes predictions on the images of the unlabeled pool. Then, uncertainty estimation is applied on the predictions and the most informative images are queried by the BALD active learning criterion, asking for annotation. If the dataset does not have background images or has few background images, background inpainting of instrument pixels in selected labeled real images is performed and the generated backgrounds are added to the background pool. Synthetic images are then generated based on the newly labeled images. For each real image, there are two types of synthetic images. Type-1 synthetic images have the same surgical tool as the original real image, and the background is randomly selected from the background pool. Type-2 synthetic images have the same background as the original real image (background inpainting is applied to the original real image to remove the original tool), and the surgical tool is randomly selected from the labeled pool. 

After the generation, the newly labeled real images and the synthetic images are added to the labeled pool and then the segmentation model is trained again to start a new iteration, until the labeling budget or the desired performance is reached. Budget in this paper is defined as the fraction of training images which are given annotation, compared to the total number of training images.

\subsection{Generation of Synthetic Image} \label{sch_gen_syn}

One synthetic image is generated from a real labeled image containing a surgical tool, and from a background image, either a real background or an inpainting background. The overall goal is to copy the tool from the instrument image and paste it on the background image, with resizing, movement and fusion. Fig. \ref{gen_syn} shows the workflow of the generation of synthetic images.

\begin{figure}
\centering
\vspace{0.5em}
\includegraphics[width=0.35\textwidth]{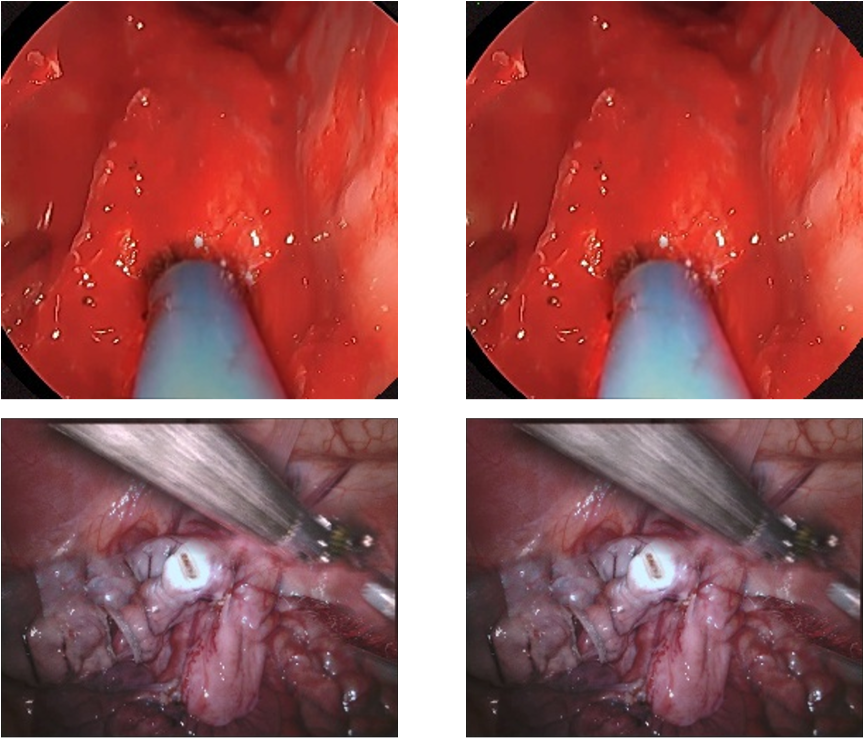}
\vspace{-1.em}
\caption{Multi-blending: images on the left and right have the same size and position of instrument and background, but have different blending method (average fusion on the left and Gaussian fusion on the right) and parameters.} 
\vspace{-1.5em}
\label{multi_blend}
\end{figure}

\begin{figure}
\centering
\vspace{0.5em}
\includegraphics[width=0.35\textwidth]{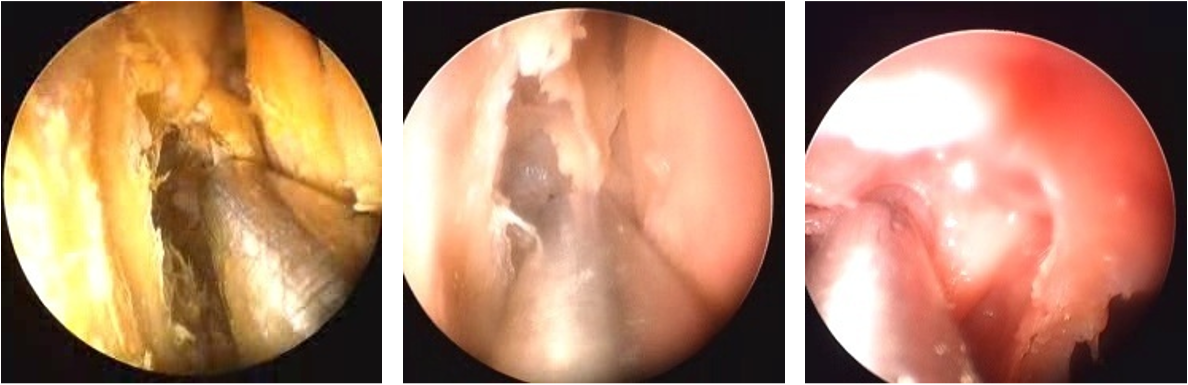}
\vspace{-1.em}
\caption{Example of images from UW-Sinus-Surgery-C/L Dataset dataset, where the outlooks of the instrument are different due to reflection.} 
\vspace{-1.5em}
\label{visual_diff}
\end{figure}

\begin{figure}
\centering
\vspace{0.5em}
\includegraphics[width=0.35\textwidth]{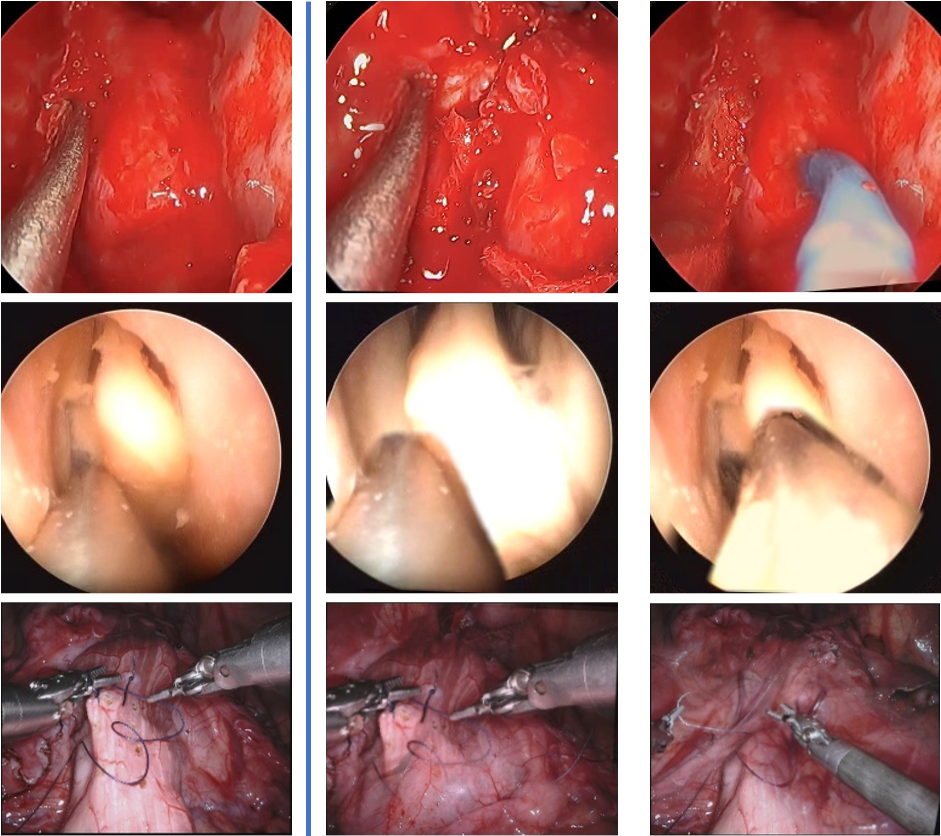}
\vspace{-1.em}
\caption{Original real images (left), Type-1 synthetic images (center) and Type-2 synthetic images (right). Type-1 has the same instrument but a different background, and type-2 has the same background but a different instrument.} 
\vspace{-1.5em}
\label{example_syn}
\end{figure}

The procedure starts with 2 images. $I_I$, a labeled real image, includes an instrument and $I_B$ is a pure background. The instrument image $I_I$ also has a mask $M_I$, which is a binary matrix with the same size as $I_I$ in which the instrument pixels are $1$ while other pixels are $0$. Resizing, movement and rotation is first applied to the instrument and the mask:
\begin{equation}
\begin{aligned}
I_{Ir}=R(I_I,c,w,h,\theta)
\end{aligned}
\end{equation}
\begin{equation}
\begin{aligned}
M_{Ir}=R(M_I,c,w,h,\theta)
\end{aligned}
\end{equation}
where $R(\cdot)$ is the operator, $c$, $w$, $h$ and $\theta$ are the factors of resizing, movement in width and height, and angle of rotation, respectively. These operations are applied sequentially. Then, a binary dilation is applied on the new mask $M_{Ir}$, so that in the dilated mask $M_{Id}$, the region of the instrument is larger than the true mask of the instrument $M_{Ir}$.
\begin{equation}
\begin{aligned}
M_{Id}=\displaystyle\bigcup_{b \in B}M_{Irb}
\end{aligned}
\end{equation}
where $B$ is a $d\times d$ matrix and $d$ is the dilation kernel size, $M_{Irb}$ is the translation of $M_{Ir}$ by $b$. After dilation, the fusion mask $M_F$ is generated by applying average blur or Gaussian blur \cite{young1995recursive} on the mask $M_{Id}$. The reason for having 2 different blurs (blending methods) and details will be introduced later in this section.
\begin{equation}
\begin{aligned}
M_F=\frac{1}{100}B_a(100M_{Id},k)
\label{average_fusion}
\end{aligned}
\end{equation}
\begin{equation}
\begin{aligned}
M_F=\frac{1}{100}B_G(100M_{Id},k,\sigma)
\label{gauss_fusion}
\end{aligned}
\end{equation}
where $B_a(\cdot)$ and $B_G(\cdot)$ are the operators of average blurring and Gaussian blurring, respectively, $k$ is the kernel size and $\sigma$ is the standard deviation. The reason for using two different fusion and blending methods is that according to \cite{dwibedi2017cut}, the artifacts from the copy-and-paste operation may result in decreased performance if the model is trained on the synthetic images. In particular, having 2 exact same synthetic images but only with different blending methods in the training set can prevent the model from learning blending artifacts and improves performance on real images, as Fig. \ref{multi_blend} shows. We study the effectiveness of this idea on our images in \ref{sch_exp2_type}. 
In the UW-Sinus-Surgery-C/L Dataset, it is visually apparent that instruments acquire coloration from the background through a diffuse reflection, as shown in Fig. \ref{visual_diff}.  To capture this effect in our synthetic images, color and brightness adjustment is applied first to narrow the gap of the color style of the instrument and the background, for each channel of the adjusted image:
\begin{equation}
\begin{aligned}
I_{Ic,chn}=\beta \frac{\sum I_B}{\sum I_{Ir}} \left( \alpha \frac{\sum I_{B,chn}}{\sum I_{Ir,chn}}I_{Ir,chn}+(1-\alpha )I_{Ir,chn} \right)
\end{aligned}
\end{equation}
where $\alpha$ is the factor of color adjustment, $\beta$ is the factor of brightness adjustment, $I_{B,chn}$ and $I_{Ir,chn}$ are the same channel of the background image and instrument image, respectively.
After the adjustment of color and brightness, the instrument is blended on the background by:
\begin{equation}
\begin{aligned}
I_F=M_F \bigodot I_{Ic} + (J-M_F) \bigodot I_B
\end{aligned}
\end{equation}
where $\bigodot$ indicates for element-wise multiplication, $J$ is a matrix of ones with the same size as $M_F$. After the instrument and the background are combined, to imitate endoscopic vision, we apply a weak Gaussian blur and trim the border to restore the outline, and thus finalize the generation of the synthetic image $I_{Syn}$ and the corresponding mask $M_{Syn}$. For UW-Sinus-Surgery-C/L Dataset:
\begin{equation}
\begin{aligned}
I_{Syn}=B_G(T_c(I_F,x_o,y_o,r),k_f,\sigma_f)
\end{aligned}
\end{equation}
\begin{equation}
\begin{aligned}
M_{Syn}=T_c(M_{Ir},x_o,y_o,r)
\end{aligned}
\end{equation}
where $T_c(\cdot)$ is the circle trimming operator, $(x_o, y_o)$ and $r$ are the center and radius of the trimming circle, respectively, and $B_G(\cdot)$  is the operator of Gaussian blur with kernel size $k_f$ and standard deviation $\sigma_f$.
Similarly, for EndoVis 2017 dataset, because the vision is rectangular:
\begin{equation}
\begin{aligned}
I_{Syn}=B_G(T_r(I_F, t_t,t_b,t_l,t_r), k_f, \sigma_f)
\end{aligned}
\end{equation}
\begin{equation}
\begin{aligned}
M_{Syn}=T_r(M_{Ir}, t_t,t_b,t_l,t_r)
\end{aligned}
\end{equation}
where $T_r(\cdot)$ is the rectangular trimming operator and $t_t$, $t_b$, $t_l$, $t_r$ are the trimming width of top, bottom, left and right, respectively. Fig. \ref{example_syn} shows some examples of Type-1 and Type-2 synthetic images.

\subsection{Inpainting of Backgrounds} \label{sch_inpaint_bkgd}
As introduced in \ref{sch_gen_syn}, a synthetic image could be generated from a labeled instrument image and a background image. However, it is not always possible to find background images in every dataset. Thus, for those datasets without backgrounds, image inpainting is performed to generate backgrounds from labeled instrument images.
\begin{figure}
\centering
\vspace{0.5em}
\includegraphics[width=0.35\textwidth]{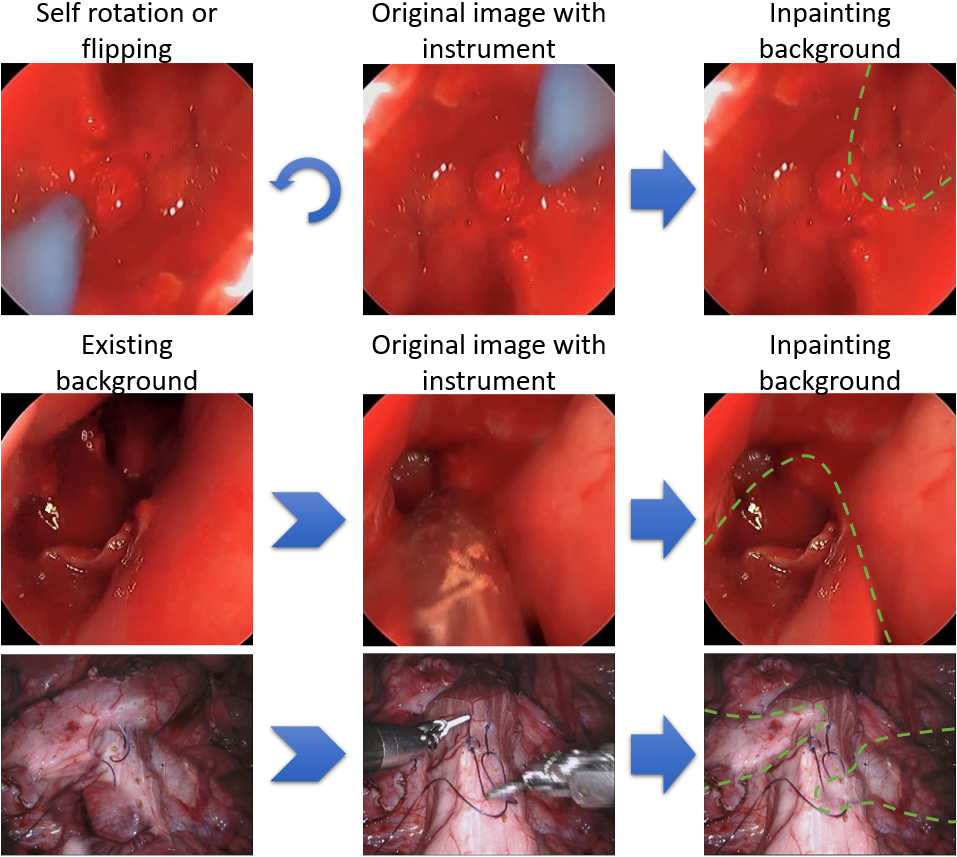}
\vspace{-1.em}
\caption{Generation of inpainting backgrounds} 
\vspace{-1.5em}
\label{inpaint_bkgd}
\end{figure}
Fig. \ref{inpaint_bkgd} shows the procedure of background inpainting. The inpainting of the background image is similar to the generation of synthetic images. The difference is that for the inpainting of background, an area of background is blended over the instrument pixels, instead of blending an instrument over a background. There are 2 types of inpainting, self-inpainting and external inpainting. First, self-inpainting can be performed by self-flipping or rotation of the original image (\ref{self-flipping1}), if the flipped or rotated mask does not overlap with the original mask (\ref{self-flipping2}):
\begin{equation}
\begin{aligned}
I_{Bi}=M_F \bigodot I_{Ir} + (J-M_F) \bigodot I_I
\label{self-flipping1}
\end{aligned}
\end{equation}
\begin{equation}
\begin{aligned}
M_F \cap M_{Fr} =0
\label{self-flipping2}
\end{aligned}
\end{equation}
where $I_I$ is the image including instrument, $I_{Bi}$ is the inpainting background with instrument removed, $M_F$ is the fusion mask generated by the method mentioned in \ref{sch_gen_syn}, $I_{Ir}$ and $M_{Fr}$ are the flipped or rotated image $I_I$ and mask $M_F$. Flipping can be applied vertically or horizontally, and rotation can be applied for degrees of $90$, $180$ and $270$.

\begin{figure}
\centering
\vspace{0.5em}
\includegraphics[width=0.35\textwidth]{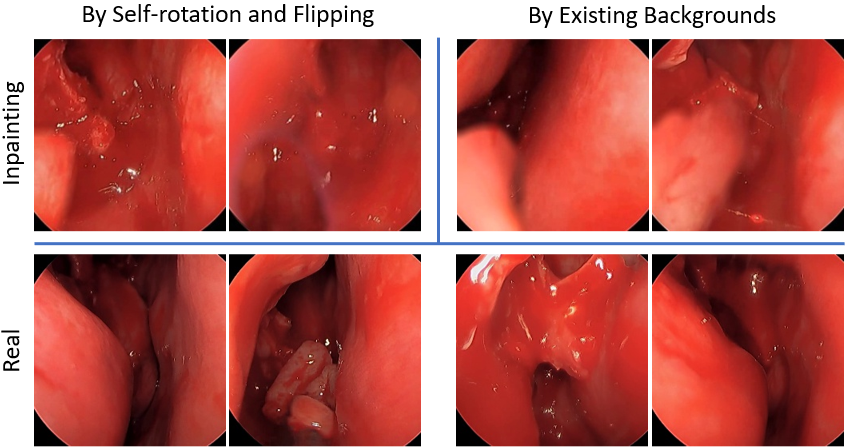}
\vspace{-1.em}
\caption{Example of inpainting backgrounds (top) and real backgrounds (bottom).} 
\vspace{-1.5em}
\label{example_inpaint_bkgd}
\end{figure}

However, sometimes the masks $M_F$ and $M_{Fr}$ always overlap regardless of flipping and rotation. In this case, external inpainting can be performed by randomly selecting another background from the background pool as a source for the pixels covering the instrument:
\begin{equation}
\begin{aligned}
I_{Bi}=M_F \bigodot I_{B2} + (J-M_F) \bigodot I_I
\end{aligned}
\end{equation}
where $I_{B2}$ is the background (original or inpainting) from the background pool and other variables are the same as self-inpainting. Fig. \ref{example_inpaint_bkgd} shows the comparison between original (real) backgrounds and inpainting (synthetic) backgrounds.

\subsection{Active Learning}
Active learning (AL) is a feasible technology to reduce the load of annotating while having competitive performance of the model trained by limited data. The proposed system uses a pool based AL method shown in Fig. \ref{workflow}, in which there are a limited number of labeled images and plenty of unlabeled images. The iteration of active learning is introduced in \ref{sch_sys_workflow} and BALD is used as the criterion to query unlabeled images.

The BALD criterion is to choose the images which are expected to maximise the mutual information between predictions and model posterior. To perform BALD, Monte-Carlo (MC) dropout, which is a frequently used stochastic regularization technique, is performed during training and inference. 
\begin{equation}
\begin{aligned}
\mathbb{I}[y,\omega | x , \mathcal{D}_{train}] = \mathbb{H}[y | x,\mathcal{D}_{train}] - \mathbb{E}_{P(\omega|\mathcal{D}_{train})} \left[\mathbb{H}[y | x,\omega] \right]
\end{aligned}
\end{equation}
where $x$ is the input image and $y$ is the output label, $\mathbb{H}[y | x,\mathcal{D}_{train}]$ and $\mathbb{H}[y | x,\omega]$ are the entropy \cite{shannon1948mathematical} of the prediction $P(\omega|\mathcal{D}_{train})$ and distribution $p(y|x, \omega)$, respectively. $\mathcal{D}_{train}$ is the labeled training data, and $\omega$ are model weights. The first term seeks the images which have high average entropy in the sampled models. And the second term gives the penalty such that the images on which the models disagree are kept and images on which many models are unconfident are dropped.  

To be more specific, implementation of the BALD criterion on image semantic segmentation is:
\begin{equation}
\begin{aligned}
\mathbb{I}[y,\omega | x , \mathcal{D}_{train}] \approx {} & - \displaystyle\sum_{c} \left(\frac{1}{T} \displaystyle\sum_{t}p_c^t \right) \log (\frac{1}{T} \displaystyle\sum_{t}p_c^t) \\
& + \frac{1}{T} \displaystyle\sum_{t,c}p_c^t \log p_c^t
\end{aligned}
\end{equation}
where $c$ is the number of classes, $T$ is the number of committee members (model trained and inferred  with MC dropout), $p^t_c$ is the softmax probability of the pixel.

\section{Experiments and Results}

\subsection{Datasets} \label{sch_datasets}
All the images in the datasets were manually labeled by experts (surgeons and surgical residents). However, ground truth masks are provided only when the images are in the test set or are marked as ‘labeled’.  The labeled images can be used for initial training or queried by AL.

\textbf{UW-Sinus-Surgery-C/L Dataset} \cite{qin2020towards} contains two parts: the live dataset (Sinus-Live) and cadaver dataset (Sinus-Cadaver). The live dataset is collected from the videos of 3 live surgeries on 3 patients. The duration of the videos is around 2.5 hours in total, with image resolution of 1920$\times$1280 and frame rate of 30 fps. 3955 images from the first two videos are used as the training set. 696 background images are generated by subsampling the videos at 3 Hz and manually selecting background-only frames. These selected backgrounds are provided to the system when external backgrounds are asked for. Manually choosing backgrounds is not as costly as manually annotating the images. Three subjects without surgical background are asked to select 700 backgrounds out of 20000 images and the time spent is 25 min, 17 min and 31 min, respectively. 703 hand-labeled images from the third video are used as the test set. All the images are resized and center cropped to 240$\times$240. 

In order to prove the ability of generalization, the images in the test set and training set are from different videos from different surgical procedures. Because there are less than 15 real images used in some of our present experiments, meaningless images (such as pure black or white images caused by over-exposure or blocking) were manually cleaned from the training set. However, the meaningless images are not cleaned in the test set to ensure that the performance is fairly evaluated.

In Cadaver dataset, the sinus surgery video dataset is built similarly, collected from 10 surgeries on 5 cadaver specimens. The training set is generated from the first 7 videos, which has 2908 images. The test set is generated from the remaining 3 videos, which has 1437 images. 597 backgrounds are chosen from the training set videos. And the images are also resized and center cropped to 240$\times$240. Due to the different condition of each cadaver specimen, the overall appearance of the images can be different. However, none of the recorded videos shows considerable similarity to real surgeries.   Humans have no difficulty to visually distinguish cadaver videos and live videos. 

\textbf{EndoVis 2017 Robotic Instrument Segmentation Dataset} is from one of the sub-challenges of MICCAI 2017 \cite{allan20192017}. The images are derived from 10 sequences of abdominal porcine procedures recorded using da Vinci Xi robotic endoscopic surgery systems, when significant instrument motion can be observed. The instruments used include Large Needle Driver, Prograsp Forceps, Monopolar Curved Scissors, Cadiere Forceps, Bipolar Forceps, Vessel Sealer and an ultrasound probe. 300 images are collected per video at 1 Hz and repetitive images caused by non-moving instruments are manually cleaned. The selected frames were labeled by a segmentation team at Intuitive Surgical. Although the videos were recorded by stereo camera, only left eye images are labeled. We used 900 images (225 each video) with labels, from videos 1-4, as our training set. 900 images and labels from videos 5-8 are used as our test set. Due to the long training time of active learning, the images are resized to 427$\times$240 to reduce computational time.

\subsection{Training Details}
There are two main groups of parameters, parameters related to the generation of synthetic images, and parameters related to model training and active learning.

The segmentation model used in this paper has the same structure as \cite{qin2020towards}. This is a modified DeepLabv3+ \cite{deeplabv3plus2018} encoder-decoder model with Mobilenet \cite{howard2017mobilenets} as feature extractor.  In order to fit in the active learning iterations,  the learning rate is increased and meanwhile the training iterations are decreased significantly to accelerate the training, and the accuracy was slightly compromised due to accelerated training speed. Adam \cite{kingma2014adam} is used as the optimizer, and the exponential decay rates of 1st and 2nd order moment estimates were 0.9 and 0.999, respectively. Batch size is set as 16. Because we use different budgets of the training set in the experiments, for all the experiments in this paper, the training iterations was set to be equivalent to 5 epochs on 100\% training set (real images) plus 20 basic epochs to ensure convergence when a higher proportion of training set is used. Initial learning rate is set as 0.001 and exponential decay strategy is applied. The backbone lightweight MobileNet is pretrained on ImageNet \cite{ILSVRC15}. Image argumentation was also applied to the training data for better generalization ability, which includes hue, brightness, saturation, contrast, flipping, rotation, zooming and zero-padding. The model is trained on a local Nvidia GTX Titan X GPU and remote NVIDIA Tesla P100 GPUs from Google Colaboratory.

We tested active learning performance by applying a variable budget of real images. To start, all the images in the dataset were in the unlabeled set and the labels were hidden. For example, if the real-image budget was 394 images (10\% of the Live dataset), then 197(half) real images are randomly chosen first, and the other half 197 images are chosen by the BALD criterion in 3 iterations. Once chosen from the unlabeled set, the real images are moved to the labeled set and their labels are revealed. And synthetic images are generated with labels naturally.

The main parameters of synthetic images generation were empirically chosen and are shown in Table 1. These parameters apply to all the tests unless stated otherwise.

%\newcolumntype{M}[1]{>{\centering\arraybackslash}m{#1}}
%\newcolumntype{N}{@{}m{0pt}@{}}
\begin{table*}[h]
\vspace{1.em}
\caption{\vspace{-0.3em} Parameters of the generation of synthetic images}
\vspace{-1.5em}
%of our cross-domain strategy and the mainstream strategy
\begin{center}
\renewcommand{\arraystretch}{2}
\begin{tabular}{|p{3cm} | p{5cm} | p{5cm} |}
\hline
Parameter                                                   & Value (Live / Cadaver / EndoVis)                      & Description                                                                                                                                     \\
\hline
(\textbf{Type-1}, \textbf{Type-2}) Synthesis per query                        & (2,0) / (2,0) / (0,1)                                 & A Type-1 synthetic image has the same instrument with the original real image. A Type-2 synthetic image has the same background with the original images. \\
\hline
Multi-blending                                              & No / No / Yes                                         & Multi-blending generates two exact same synthetic images with only the difference of the blending method, to minimize the effect of artifact.             \\
\hline
External Backgrounds                                        & Yes / Yes / No                                        &                                                                                                                                                           \\
\hline
Background Inpainting                                       & No / No / Yes                                         &                                                                                                                                                           \\
\hline
Factor of resizing $c$                                        & {[}0.9, 1.2{]}                                        & The ratio of new size to the origin                                                                                                                       \\
\hline
Movement in width $w$                                         & {[}-0.1, 0.1{]} / {[}-0.1, 0.1{]} / {[}-0.05, 0.05{]} & $w$, and $h$ are the proportion of the movement. For example w=0.1 of an image with width of 100 pixels results in movement of 10 pixels.                     \\
\hline
Movement in height $h$                                        & {[}-0.1, 0.1{]} / {[}-0.1, 0.1{]} / {[}-0.05. 0.05{]} &                                                                                                                                                           \\
\hline
Angle of rotation $\theta$                                          & {[}-30, 30{]}                                         & In degree                                                                                                                                                 \\
\hline
Dilation kernel size $d$                                     & 15 / 15 / 15                                          & Used to build fusion mask                                                                                                                                 \\
\hline
Fusion blur kernel size $k$                                   & {[}10,15{]} / {[}5, 10{]} / {[}10, 15{]}              & Used to build fusion mask                                                                                                                                 \\
\hline
Factor of color adjustment $\alpha$                                 & {[}0.4, 1.0{]}                                        & 1.0 means that the color of the instrument from the original image is not adjusted. Smaller  means stronger adjustment                                    \\
\hline
Factor of brightness adjustment   $\beta$                          & {[}0.9, 1.3{]}                                        & The larger the brighter and 1.0 means that the brightness is not adjusted                                                                                 \\
\hline
Center of circle trimming ($x_o$, $y_o$)                      & ({[}115,125{]}, {[}115,125{]})                        & In pixels                                                                                                                                                 \\
\hline
Radius of circle trimming $r$                                 & {[}150, 170{]}                                        & In pixels                                                                                                                                                 \\
\hline
Width of rectangular trimming ($t_t$, $t_b$, $t_l$, $t_r$)              & ({[}6, 9{]}, {[}6, 9{]}, {[}71, 74{]}, {[}71, 74{]})  & In pixels                                                                                                                                                 \\
\hline
Kernel size and standard deviation of Gaussian blur ($k_f$, $\sigma_f$) & (3, 3)                                                &   \\
\hline
\multicolumn{3}{l}{* [a, b] means that the parameter is randomly chosen from range a to b.}\\
\end{tabular}
\end{center}
\vspace{-2.5em}
\label{tab_param}
\end{table*}

\subsection{Evaluation Metrics}
Two main evaluation metrics are used in this paper, Dice similarity coefficient (DSC) and intersection of union (IoU) \cite{taha2015metrics}, which are defined as
$$
DSC = \frac{2|S\cap G|}{|S|+|G|},IoU = \frac{|S\cap G|}{|S\cup G|}
$$
where $S$ is the foreground pixels of prediction, $G$ is the corresponding ground truth, and $|*|$ is the counting operation. To study the effect of blending and fusion on the performance of segmentation near boundary, IoU near boundary (IoU\textsubscript{NB}) is also used as an additional metric:
$$
IoU_{NB} = \frac{|S\cap G| \cap B}{|S\cup G| \cap B}
$$
where $B$ denotes the near-boundary binary mask with width of 20 pixels band region near the instruments’ boundary. The mean values of these three metrics are calculated over each test, denoted as mDSC, mIOU and mIoU\textsubscript{NB}.

\subsection{Experiment 1: Usage of real images}
We compare the proposed method with baseline results, which are obtained when 100$\%$ of real images in the training set are labeled and used to train the segmentation model. Additionally we study the performance when different annotation budgets are used. Budgets are set as proportions of the total real images in the training set, from 1$\%$ to 100$\%$. For example, 1$\%$(39), 5$\%$(197), 10$\%$(395) images were used from the Sinus-Live training dataset (3955 images in total), and 1$\%$(29), 5$\%$(145), 10$\%$(290) images are used from the Sinus-Cadaver training dataset (2908 images in total). However, the training on EndoVis 2017 dataset did not converge on the 1$\%$ or 2$\%$ budget because of too few images. Thus, the evaluation of this dataset started at 5$\%$. For each budget, 4 tests were performed - 1) randomly chosen training images without synthetic images, 2) BALD implemented - half chosen by BALD and the other half chosen randomly, without synthetic images, 3) randomly chosen images with generated synthetic images, 4) BALD implemented with generated synthetic images.   

\begin{figure*}
\centering
\vspace{0.3em}
\includegraphics[width=0.7\textwidth]{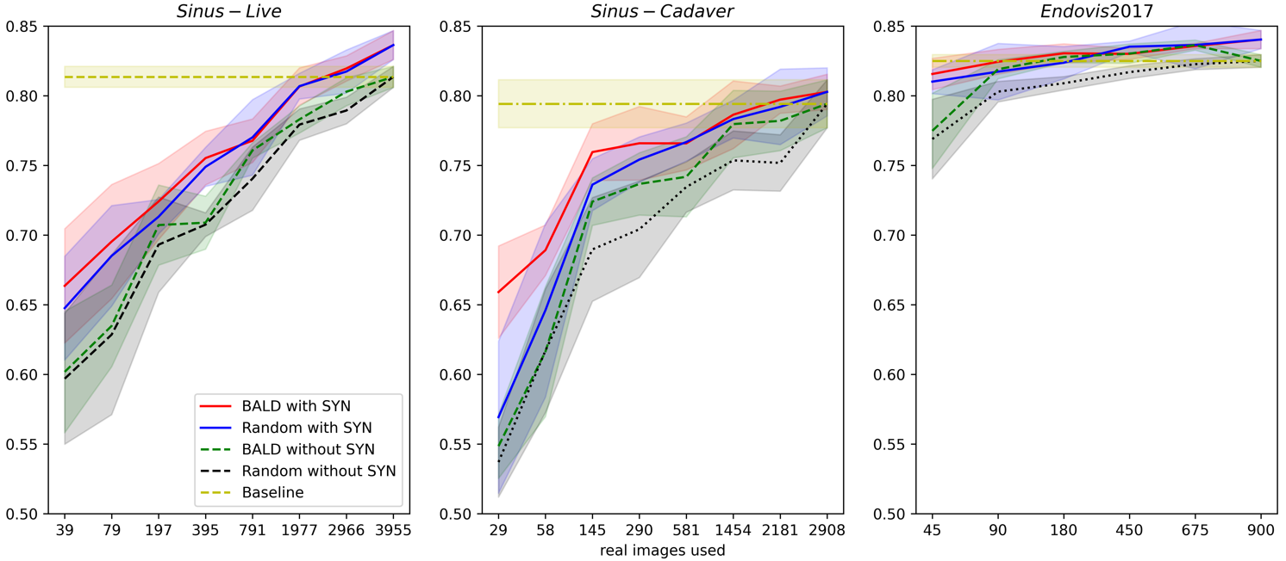}
\vspace{-1.em}
\caption{Evaluation (mDSC) of the model with different budgets of real images, with and without active learning (BALD) and synthetic images (SYN). Please notice that the horizontal axises are nonlinear.} 
\vspace{-1.5em}
\label{rst_exp1_real_img}
\end{figure*}

The evaluation results of mDSC are given in Fig. \ref{rst_exp1_real_img} , and more details of mIoU and mIoUnb are shown in Table 4 in Appendix. Each entry is the average of 5 repetitive tests with different global random seeds to reduce the influence of randomness, which applies to all the experiments in this paper. From Fig. \ref{rst_exp1_real_img}, it can be seen that by generating synthetic images with active learning, the performance of segmentation is significantly improved when the number of real labeled images (budget) is small. Compared with randomly chosen and no synthetic images, the average improvement  of the proposed method in mDSC with small budgets (less than 10$\%$ of training set) is 5.31$\%$, 8.15$\%$ and 3.41$\%$ in Sinus-Live, Sinus-Cadaver and EndoVis 2017 dataset, respectively. Specifically, the improvements on the Sinus-Live and Sinus-Cadaver dataset with only 1$\%$ budget are 6.67$\%$ and 12.20$\%$. Overall, as the labeled image budget increases, the improvement becomes smaller. However, a considerable improvement can still be achieved when the budget is 100$\%$ of the training set and BALD active learning is not applicable. When the budget is 100$\%$, generation of additional synthetic images still results in an improvement of 2.29$\%$, 0.86$\%$ and 1.54$\%$ in the three datasets, respectively. Compared with the results on the sinus datasets,  the average improvement in EndoVis 2017 dataset is not as significant, but the result of the proposed method begins to outperform the baseline result at only 10$\%$ usage of the training set.

From Table 4 in Appendix, similar trends can be seen in mIoU and mIoU near the boundary. The average improvements of the proposed method with small budgets (less than 10$\%$) are significant in mIoU - 7.11$\%$, 10.02$\%$ and 4.93$\%$ in the three datasets, as well as mIoU\textsubscript{NB} - 5.22$\%$, 7.42$\%$ and 6.39$\%$. Considerable improvements in mIoU can still be seen when the budget is 100$\%$ of the training set - 3.04$\%$, 1.33$\%$ and 2.14$\%$. However, for mIoU\textsubscript{NB}, only the result of the Sinus-Live dataset shows improvement (4.52$\%$), while the results of the other two datasets are close to the baseline results.

\subsection{Experiment 2: Number and type of synthetic images} \label{sch_exp2_type}
As introduced in \ref{sch_gen_syn}, there are 2 types of synthetic images. For each chosen and labeled real image, Type-1 synthetic images have the same instrument and Type-2 synthetic images have the same background as the real image. And multi-blending is also reported in \cite{dwibedi2017cut} to be able to avoid decreased performance caused by the artifact near the boundary in synthetic images. Thus, this experiment is performed to study the effectiveness of the 2 types of synthetic images and multi-blending.  To better compare the results, the experiments are separated into 3 groups. Each group tests the same number of generated images per BALD query. For example, in Table Table \ref{Tab_syn}, tests in Group 2 feature 8 synthetic images for each queried real image - $(2[Type-1] + 2[Type-2]) \times 2[Multi-blending]$.   Consequently, tests in each group have the same number of training iterations to ensure that the model is trained for the same fixed number of steps in each case. To ensure convergence of training, instead of keeping training iterations , the training epochs of Group 2 and 3 are the same as Group 1 so that the training iterations of Group 2 and 3 are four and six times larger compared to Group 1, respectively. All the other parameters are set as Table 1 and the annotation budget is fixed at 10$\%$ of the training set. 

The results are shown in Table \ref{Tab_syn}. The parameter of Type-1 and Type-2 means that for each labeled real image selected by the active learning mechanism, how many Type-1 and Type-2 synthetic images were generated, respectively. A “multi-blending” value of 1 means that each synthetic image is single and multi-blending is not applied. And multi-blending value of 2 means that for each synthetic image blended by average fusion (\ref{average_fusion}), there is another similar synthetic image blended by Gaussian fusion (\ref{gauss_fusion}). In Group 1, for the two sinus surgery datasets, the best performance on mDSC and mIoU were achieved by 2 Type-1 synthetic images, while the best performance on mIoU\textsubscript{NB} was achieved by 1 Type-2 synthetic image with multi-blending on Sinus-Live dataset, and by 1 Type-1 synthetic image with multi-blending on Sinus-Cadaver dataset. And for the EndoVis 2017 dataset, 1 Type-2 synthetic image with multi-blending gives the best result on mDSC and mIoU, and 1 Type-1 synthetic image with multi-blending gives the best result on mIoU. 

In Group 2, although the training steps are increased significantly compared to group 1, a decrease in performance can be seen in the Sinus-Cadaver and EndoVis 2017 datasets compared to Group 1.  Within the group, there is no considerable difference in results with and without multi-blending on the Sinus-Live and EndoVis 2017 dataset. However, multi-blending increases the performance significantly in the Sinus-Cadaver dataset. Similar trend can be observed in Group 3. Although no considerable difference is seen on mDSC and mIoU in Sinus-Live and EndoVis 2017 dataset, multi-blending gives an improvement on mIoU\textsubscript{NB} of around 1$\%$.

\begin{table*}[h]
\vspace{1.em}
\caption{\vspace{-0.3em} Segmentation Result with Different Types of Synthetic Images and Multi-blending}
\vspace{-1.5em}
%of our cross-domain strategy and the mainstream strategy
\begin{center}
\begin{tabular}{c | c : c : c | c  c  c | c  c  c | c  c  c}
\hline
\multirow{3}{*}{Group} & \multicolumn{3}{c|}{\multirow{2}{*}{Syn per Real Image}} & \multicolumn{9}{c}{Performance(\%)}                                                  \\
\cline{5-13}
                         & \multicolumn{3}{l|}{}                                    & \multicolumn{3}{c|}{Live} & \multicolumn{3}{c|}{Cadaver} & \multicolumn{3}{c}{EndoVis} \\
                         \cline{2-13}
                         & Type-1           & Type-2           & M-blend           & mDSC   & mIOU   & mIoU\textsubscript{NB} & mDSC    & mIOU    & mIoU\textsubscript{NB}  & mDSC    & mIOU    & mIoU\textsubscript{NB}  \\
\hline
\hline
\multirow{7}{*}{1}       & \multicolumn{3}{c|}{No synthetic image}                  & 71.70  & 64.19  & 55.47  & 69.35   & 61.81   & 52.76   & 79.55   & 68.06   & 67.88   \\
                         & 1                & 1                & 1                 & 74.74  & 68.24  & 59.34  & 73.32   & 66.89   & 56.15   & 81.94   & 71.77   & 71.76   \\
                         & 0.5              & 0.5              & 2                 & 74.64  & 68.09  & 60.54  & 74.30   & 67.80   & 58.31   & 81.47   & 71.19   & 71.14   \\
                         & 2                & 0                & 1                 & \textbf{76.97}  & \textbf{70.47}  & 61.65  & \textbf{76.59}   & \textbf{69.93}   & 59.22   & 81.37   & 71.14   & 72.62   \\
                         & 1                & 0                & 2                 & 74.75  & 68.50  & 60.16  & 75.29   & 68.45   & \textbf{59.24}   & 81.34   & 71.35   & \textbf{74.24}   \\
                         & 0                & 2                & 1                 & 75.07  & 68.67  & 60.63  & 71.90   & 65.77   & 55.69   & 82.25   & 72.03   & 71.60   \\
                         & 0                & 1                & 2                 & 76.80  & 70.38  & \textbf{63.23}  & 72.85   & 66.44   & 56.34   & \textbf{82.43}   & \textbf{72.46}   & 73.31   \\
\hline
\multirow{2}{*}{2}       & 4                & 4                & 1                 & \textbf{78.03}  & \textbf{72.23}  & \textbf{62.96}  & 70.86   & 65.33   & 54.84   & 80.90   & \textbf{70.48}   & 68.50   \\
                         & 2                & 2                & 2                 & 77.57  & 71.89  & 62.91  & \textbf{76.09}   & \textbf{70.32}   & \textbf{59.67}   & \textbf{80.96}   & 70.39   & \textbf{68.57}   \\
\hline
\multirow{2}{*}{3}      & 6                & 6                & 1                 & 77.42  & 71.73  & 61.95  & 69.90   & 64.46   & 54.21   & 80.43   & 69.77   & 66.96   \\
                         & 3                & 3                & 2                 & \textbf{77.91}  & \textbf{72.15}  & \textbf{62.79}  & \textbf{74.90}   & \textbf{69.45}   & \textbf{58.91}   & \textbf{80.95}   & \textbf{70.29}   & \textbf{67.61}  \\
\hline
\multicolumn{13}{l}{*The bold font indicates the best performance in each group. The parameter of Type-1 and Type-2 means that for each labeled real image } \\
\multicolumn{13}{l}{selected by the active learning mechanism, how many Type-1 and Type-2 synthetic images were generated, respectively. A M-blend value} \\
\multicolumn{13}{l}{of 1 means that each synthetic image is single and multi-blending is not applied. And M-blend value of 2 means that for each synthetic} \\
\multicolumn{13}{l}{image blended by average fusion (\ref{average_fusion}), there is another similar synthetic image blended by Gaussian fusion (\ref{gauss_fusion}).}
\end{tabular}
\end{center}
\vspace{-2.5em}
\label{Tab_syn}
\end{table*}

\subsection{Experiment 3: Strength of Fusion and Blending}
\begin{figure}
\centering
\vspace{0.5em}
\includegraphics[width=0.35\textwidth]{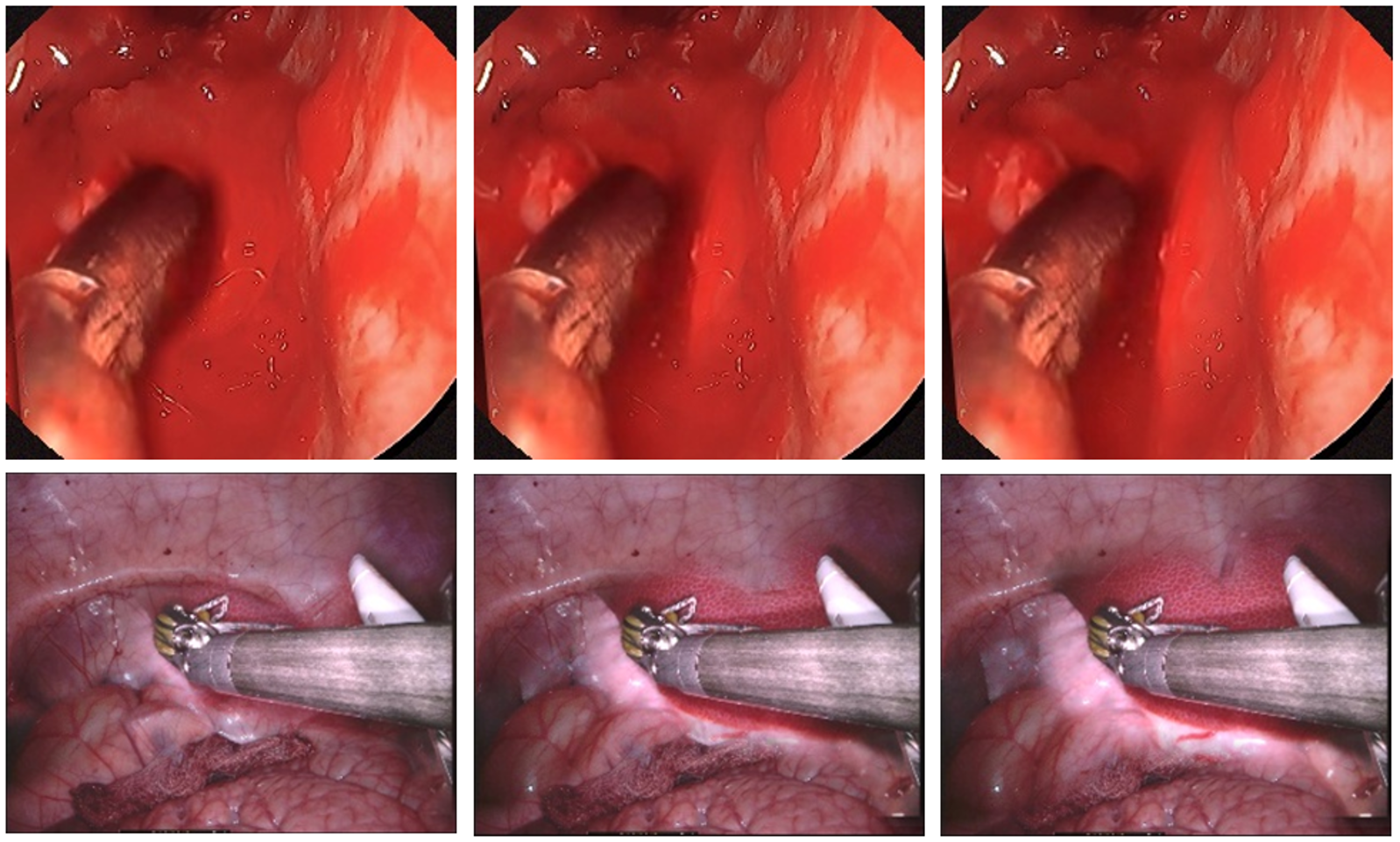}
\vspace{-1.em}
\caption{Synthetic images generated from weak fusion $d$=15, $k$=10 (left, less instrument background is retained), medium fusion $d$=40, $k$=30 (middle) and strong fusion $d$=60, $k$=45 (right, more instrument background is retained).} 
\vspace{-1.5em}
\label{example_fusion_diff}
\end{figure}

As introduced in \ref{sch_gen_syn},  the fusion and blending of borders synthetic instrument and background of synthetic images are controlled by 2 parameters, dilation kernel size $d$ and fusion kernel size $k$. Generally speaking, a larger $d$ results in a larger area near the instrument in the original real image blended on the new synthetic image. And a larger $k$ results in a larger transition area near the fusion borders. Fig. \ref{example_fusion_diff} shows an example of 3 images from weak fusion, medium fusion and strong fusion.

Thus, to study the effectiveness of fusion and blending, different combinations of d and k are  evaluated while all the other parameters are held fixed according to Table 1. The result is shown in Table 3. Group 0, 1, 2, and 3 are no fusion, weak fusion, medium fusion and strong fusion, respectively. The difference in performance is around 1$\%$-3$\%$, not as large as that in the previous test. It can be seen that most of the best performances are found in group 1. However, for Sinus-Live and EndoVis 2017 dataset, the best results in mIoU\textsubscript{NB} is found in group 3.

\begin{table*}[h]
\vspace{1.em}
\caption{\vspace{-0.3em} Segmentation Result with Different Fusion and Blending}
\vspace{-1.5em}
%of our cross-domain strategy and the mainstream strategy
\begin{center}
\begin{tabular}{ c | c : c | c  c  c | c  c  c | c  c  c}
\hline
\multirow{3}{*}{Group} & \multicolumn{2}{c|}{\multirow{2}{*}{Proposed}} & \multicolumn{9}{c}{Performance(\%)}                                                  \\
\cline{4-12}
                       & \multicolumn{2}{c|}{}                            & \multicolumn{3}{c|}{Live} & \multicolumn{3}{c|}{Cadaver} & \multicolumn{3}{c}{EndoVis} \\
\cline{2-12}
                       & Dilation Kernel $d$       & Fusion Kernel $k$       & mDSC   & mIoU   & mIoU\textsubscript{NB} & mDSC    & mIoU    & mIoU\textsubscript{NB}  & mDSC    & mIoU    & mIoU\textsubscript{NB}  \\
\hline
\hline
0                      & \multicolumn{2}{c|}{No fusion}                   & 74.03  & 67.39  & 59.27  & 72.77   & 65.97   & 57.38   & 82.31   & 72.26   & 72.98   \\
\hline
\multirow{3}{*}{1}     & 15                      & {[}5, 10{]}            & 73.85  & 66.94  & 59.83  & \textbf{74.90}   & \textbf{68.16}   & \textbf{58.02}   & \textbf{82.82}   & 72.81   & 72.26   \\
                       & 15                      & {[}10, 15{]}           & \textbf{74.96}  & \textbf{68.44}  & 59.87  & 74.48   & 67.59   & 57.39   & 82.81   & \textbf{72.94}   & 73.25   \\
                       & 15                      & {[}15, 20{]}           & 74.75  & 67.88  & 58.34  & 74.69   & 67.78   & 57.58   & 82.24   & 72.11   & 72.16   \\
\hline
\multirow{3}{*}{2}     & 40                      & {[}20, 30{]}           & 74.54  & 66.68  & 58.78  & 72.96   & 65.45   & 55.73   & 82.55   & 72.53   & 72.81   \\
                       & 40                      & {[}30, 40{]}           & 74.11  & 67.12  & 59.93  & 72.59   & 65.44   & 56.77   & 82.71   & 72.76   & 72.93   \\
                       & 40                      & {[}40, 50{]}           & 73.13  & 66.16  & 57.30  & 74.45   & 67.46   & 57.62   & 81.98   & 71.71   & 71.46   \\
\hline
\multirow{3}{*}{3}     & 60                      & {[}30, 45{]}           & 74.89  & 68.09  & \textbf{60.09}  & 73.82   & 66.79   & 57.53   & 82.33   & 72.34   & 72.36   \\
                       & 60                      & {[}45, 60{]}           & 74.71  & 67.65  & 59.11  & 72.82   & 65.64   & 56.03   & 82.63   & 72.55   & 72.51   \\
                       & 60                      & {[}60, 75{]}           & 74.72  & 67.50  & 58.48  & 72.12   & 65.50   & 57.28   & 82.06   & 71.90   & \textbf{73.30}  \\
\hline
\multicolumn{12}{l}{* 1)The bold font indicates the best performance in the column. 2) [a, b] means that the parameter is randomly chosen from range a to b.}\\
\end{tabular}
\end{center}
\vspace{-2.5em}
\label{Tab_fusion}
\end{table*}

\subsection{Experiment 4: External backgrounds}
For the proposed method, backgrounds are critical in generating synthetic images which we generated two ways. First we used image inpainting (\ref{sch_inpaint_bkgd}), and  second we manually identified“external” background images (without any instruments present) from the video frames in the database. This experiment studied whether providing these external backgrounds can help with the segmentation result. The tests were separated into 4 groups according to  the budget of real images (1$\%$ means the budget of the real images is equivalent to 1$\%$ of the training set), as shown in Table \ref{Tab_ext_bkgd} in Appendix. In each group, there are 4 sub-tests. The baseline test is to only use the real image to train the segmentation model without synthetic images. The remaining 3 sub-test are all with synthetic images (BALD implemented). The only difference is how the backgrounds are provided. For ‘No-Yes’ tests, all the backgrounds used to generate synthetic images are inpainting backgrounds from real images with instruments. For ‘Yes-No’ tests,  all backgrounds are external real background images and no inpainting backgrounds are generated. For ‘Yes-Yes’ tests, both external backgrounds and generated inpainting backgrounds are provided. All the other parameters were set according to Table \ref{tab_param}. Because no external backgrounds (frames without instruments) could be found in the EndoVis 2017 dataset, only Sinus-Live and Sinus-Cadaver datasets were used to perform this experiment.

The results in Fig. \ref{rst_ext_bkgd} (mDSC) and Table \ref{Tab_ext_bkgd} in Appendix show that including external backgrounds improves performance significantly when an extremely small amount of real images were used (1$\%$ of the training set). For the Sinus-Live dataset, compared with no external backgrounds, the improvement of best result (with or without inpainting backgrounds) is 2.77$\%$, 2.89$\%$ and 1.93$\%$ in mDSC, mIOU and mIoU\textsubscript{NB}, respectively. For the Sinus-Cadaver dataset, the improvement is 4.02$\%$ (mDSC), 4.86$\%$(mIOU) and 1.74$\%$(mIoU\textsubscript{NB}). However, when a large amount of real images were used, the improvement is less considerable. For the Sinus-Live dataset, when 100$\%$ of the labeled real-image training set was used, the improvement is 0.52$\%$(mDSC), 0.70$\%$(mIOU) and 0.60$\%$(mIoU\textsubscript{NB}). But for the Sinus-Cadaver dataset, although improvement can still be seen in mDSC and mIOU with external backgrounds, a decrease of performance was found in mIoU\textsubscript{NB}(-0.99$\%$).

\begin{figure}
\centering
\vspace{0.5em}
\includegraphics[width=0.45\textwidth]{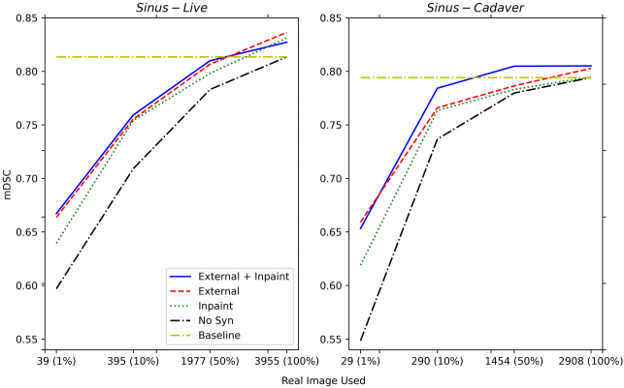}
\vspace{-1.em}
\caption{Segmentation Result (mDSC) with Different Backgrounds} 
\vspace{-1.5em}
\label{rst_ext_bkgd}
\end{figure}

%%%%%%%%%%%%%%%%%%%%%%%%%%%%%%%%%%%%%%%%%%%%%%%%%%%%%%
\section{Discussion}
In this paper, we study the use of  selectively  generated synthetic images to improve the performance of surgical instrument segmentation. The proposed method can also be easily generalized to object localization and classification. Although in this paper the effectiveness of the proposed method is validated on surgical scenes, we believe that it can also be applied to other cases such as visual object detection in self-driving cars.

The result of Experiment 1 indicates that the proposed method improves the performance of segmentation significantly, especially when few real labeled images are used. With 10$\%$ real images budget combined with active generated synthetic images, the performance of segmentation is comparable to using 50$\%$ of real images without synthetic images, cutting manual labeling effort by 80$\%$.  The performance of  the proposed method is comparable with the baseline result (using 100$\%$ of the hand labeled training set) when using only 50$\%$, 75$\%$ and 10$\%$ of hand-labeled data when evaluated on the Sinus-Live, Sinus-Cadaver and EndoVis 2017 datasets, respectively.

For sinus surgery datasets, the Type-1 synthetic images (having the same instrument as the real image but with different background) slightly outperformed Type-2 synthetic images (having different instruments combined with the same background). Although use of two different image blending methods (“multi-blending”) made no major improvement on overall performance, it does improve the segmentation near the boundary. However, increasing the number of synthetic images did not always help. Too many synthetic images can cause a decrease in the performance.

When combining (fusing) instrument and background, details such as the radius of a blend applied along the instrument boundary can have substantial effect on segmentation training.   Among different fusion strengths (Table \ref{Tab_fusion}), weaker fusion (a small area around the instrument is blended on the synthetic images but with a steep transition) gave slightly better results. The subjective realism of synthetic images for humans varied from image to image, but medium fusion and strong fusion synthetic images appear to be harder to distinguish from real images due to weaker artifacts near the border. 

We varied the method of generating background images between inpainting the space occupied by instruments, and using video frames which contained no instruments originally (“external backgrounds”). Using external backgrounds improved the performance significantly when the annotating budget was extremely small, without adding much workload. However, as the budget increased, no considerable difference was observed by using external backgrounds. Because manually selecting background-only frames from surgical videos requires no expertise and is not as time consuming as labeling the instruments, especially when different parts of the instruments have different labels, adding external backgrounds is an efficient way to improve segmentation with small numbers of labeled images.

\section{Conclusions and Future Work}
Motivated by alleviating the experts’ workload of annotating for challenging instrument segmentation in endoscopic images, we propose use of actively generated synthetic images to reduce the need for labeled real images while having comparable performance. The idea of active generated synthetic images is to select the most informative unlabeled images, then annotate these images and generate synthetic images derived from the selected real images. Thus, a more diverse training set is formed by labeled real images and synthetic images, which results in considerable improvement in performance compared with using real images only, especially when the budget for annotating “new” images is small. To sum up, the proposed method utilizes and combines active learning and generation of synthetic images to reduce the usage of real labeled images, and can be flexibly applied to different segmentation models and datasets, with different active learning criteria.

In the future, we plan  to study in principle how synthetic images help with performance. We also hope to explore the relationship between realism to humans and effectiveness to artificial intelligence, in other words, to study whether the most subjectively realistic synthetic images give the best performance on training segmentation models.

\bibliographystyle{IEEEtran}
\bibliography{IEEEabrv,IEEEexample}

%%%%%%%%%%%%%%%%%%%%%%%%%%%%%%%%%%%%%%%%%%%%%%%%%%%%%%%%%%%%%%%%%%%%%%%%%%%%%%%%%%%%%%%%%%%%

\section{Appendix}

\begin{table*}[h]
\vspace{1.em}
\caption{\vspace{-0.3em} Segmentation Performance with Different Budgets}
\vspace{-1.5em}
%of our cross-domain strategy and the mainstream strategy
\begin{center}
\begin{tabular}{ c | c : c | c  c  c | c  c  c | c  c  c}
\hline
\multirow{3}{*}{Budget} & \multicolumn{2}{c|}{\multirow{2}{*}{Proposed}} & \multicolumn{9}{c}{Performance(\%)}                                                                   \\
\cline{4-12}
                        & \multicolumn{2}{c|}{}                          & \multicolumn{3}{c|}{Sinus-Live} & \multicolumn{3}{c|}{Sinus-Cadaver} & \multicolumn{3}{c}{EndoVis 2017} \\
\cline{2-12}

                        & BALD                   & { Syn }                  & mDSC     & mIoU     & mIoU\textsubscript{NB}   & mDSC      & mIoU      & mIoU\textsubscript{NB}    & mDSC      & mIoU     & mIoU\textsubscript{NB}    \\
\hline
\hline
\multirow{4}{*}{1\%}    & $\times$                     & $\times$                   & 59.69    & 50.99    & 46.37    & 53.71     & 43.57     & 40.18     &           &          &           \\
                        & $\checkmark$                    & $\times$                   & 60.18    & 52.06    & 46.65    & 54.85     & 45.34     & 42.12     &           &          &           \\
                        & $\times$                     & $\checkmark$                  & 64.76    & 57.39    & 49.72    & 56.93     & 49.37     & 43.40     &           &          &           \\
                        & $\checkmark$                    & $\checkmark$                  & \textbf{66.36}    & \textbf{59.52}    & \textbf{53.26}    & \textbf{65.91}     & \textbf{58.51}     & \textbf{50.82}     &           &          &           \\
\hline
\multirow{4}{*}{2\%}    & $\times$                     & $\times$                   & 62.86    & 54.00    & 50.02    & 61.70     & 52.53     & 46.48     &           &          &           \\
                        & $\checkmark$                    & $\times$                   & 63.48    & 54.95    & 50.66    & 61.64     & 55.03     & 47.94     &           &          &           \\
                        & $\times$                     & $\checkmark$                  & 68.51    & 61.74    & 54.55    & 64.61     & 57.54     & 49.89     &           &          &           \\
                        & $\checkmark$                    & $\checkmark$                  & \textbf{69.54}    & \textbf{62.59}    & \textbf{54.57}    & \textbf{68.92}     & \textbf{61.82}     & \textbf{53.25}     &           &          &           \\
\hline
\multirow{4}{*}{5\%}    & $\times$                     & $\times$                   & 69.31    & 61.83    & 54.00    & 68.97     & 60.57     & 53.04     & 76.89     & 64.50    & 62.44     \\
                        & $\checkmark$                    & $\times$                   & 70.72    & 63.07    & 57.42    & 72.40     & 64.33     & 56.18     & 77.48     & 65.30    & 63.72     \\
                        & $\times$                     & $\checkmark$                  & 71.32    & 64.53    & \textbf{57.09}    & 73.62     & 66.58     & 57.57     & 81.01     & 70.73    & 70.58     \\
                        & $\checkmark$                    & $\checkmark$                  & \textbf{72.41}    & \textbf{65.86}    & 56.96    & \textbf{75.96}     & \textbf{69.12}     & \textbf{59.09}     & \textbf{81.56}     & \textbf{71.09}    & \textbf{71.28}     \\
\hline
\multirow{4}{*}{10\%}   & $\times$                     & $\times$                   & 70.75    & 63.19    & 55.18    & 70.41     & 62.63     & 53.02     & 80.29     & 69.19    & 69.37     \\
                        & $\checkmark$                    & $\times$                   & 70.89    & 63.18    & 54.85    & 73.67     & 66.04     & 56.52     & 81.90     & 71.63    & \textbf{73.47}     \\
                        & $\times$                     & $\checkmark$                  & 74.90    & 68.15    & 60.38    & 75.42     & 68.81     & 58.86     & 81.73     & 71.54    & 70.71     \\
                        & $\checkmark$                    & $\checkmark$                  & \textbf{75.53}    & \textbf{70.47}    & \textbf{61.65}    & \textbf{76.59}     & \textbf{69.93}     & \textbf{59.22}     & \textbf{82.44}     & \textbf{72.46}    & 73.31     \\
\hline
\multirow{4}{*}{20\%}   & $\times$                     & $\times$                   & 74.05    & 66.59    & 58.19    & 73.45     & 65.70     & 55.58     & 80.90     & 70.10    & 71.16     \\
                        & $\checkmark$                    & $\times$                   & 76.08    & 68.89    & 61.70    & 74.18     & 66.72     & 57.67     & 82.78     & 72.86    & \textbf{74.39}     \\
                        & $\times$                     & $\checkmark$                  & \textbf{77.01}    & 70.19    & \textbf{62.00}    & \textbf{76.68}     & 69.61     & 59.85     & 82.37     & 72.22    & 71.41     \\
                        & $\checkmark$                    & $\checkmark$                  & 76.78    & \textbf{70.43}    & 61.82    & 76.58     & \textbf{70.27}     & \textbf{60.73}     & \textbf{83.05}     & \textbf{73.25}    & 73.20     \\
\hline
\multirow{4}{*}{50\%}   & $\times$                     & $\times$                   & 77.93    & 71.26    & 62.83    & 75.36     & 68.31     & 58.18     & 81.70     & 71.04    & 71.40     \\
                        & $\checkmark$                    & $\times$                   & 78.30    & 71.73    & 64.28    & 77.97     & 70.97     & 61.17     & 83.02     & 73.13    & \textbf{74.52}     \\
                        & $\times$                     & $\checkmark$                  & \textbf{80.69}    & \textbf{74.89}    & 66.22    & 78.33     & 71.89     & 61.33     & \textbf{83.52}     & \textbf{73.80}    & 73.08     \\
                        & $\checkmark$                    & $\checkmark$                  & 80.66    & 74.87    & \textbf{66.30}    & \textbf{78.64}     & \textbf{72.49}     & \textbf{62.49}     & 83.01     & 73.32    & 73.45     \\
\hline
\multirow{4}{*}{75\%}   & $\times$                     & $\times$                   & 78.94    & 72.47    & 64.42    & 75.18     & 67.94     & 58.23     & 82.27     & 72.08    & 73.72     \\
                        & $\checkmark$                    & $\times$                   & 80.29    & 73.93    & 65.49    & 78.20     & 71.26     & 61.97     & \textbf{83.64}     & \textbf{74.21}    & \textbf{76.91}     \\
                        & $\times$                     & $\checkmark$                  & 81.74    & 76.11    & \textbf{68.62}    & 79.20     & 73.01     & 63.02     & \textbf{83.64}     & 74.17    & 73.58     \\
                        & $\checkmark$                    & $\checkmark$                  & \textbf{81.96}    & \textbf{76.24}    & 67.30    & \textbf{79.72}     & \textbf{73.68}     & \textbf{63.31}     & 83.57     & 74.04    & 74.04     \\
\hline
\multirow{2}{*}{100\%}  & $\times$                     & $\times$                   & 81.35    & 75.14    & 66.34    & 79.42     & 72.50     & 62.61     & 82.50     & 72.52    & \textbf{74.28}     \\
                        & $\times$                     & $\checkmark$                  & \textbf{83.64}    & \textbf{78.18}    & \textbf{70.86}    & \textbf{80.28}     & \textbf{73.83}     & \textbf{62.84}     & \textbf{84.04}     & \textbf{74.66}    & 74.21    \\
\hline
\multicolumn{12}{l}{*The training on EndoVis 2017 database can not converge when using 1$\%$ and 2$\%$ of the training set, due to very few images.}\\
\multicolumn{12}{l}{The bold font indicates the best performance in each budget.}
\end{tabular}
\end{center}
\vspace{-2.5em}
\label{Tab_seg}
\end{table*}

\begin{table*}[h]
\vspace{1.em}
\caption{\vspace{-0.3em} Segmentation Result with Different Backgrounds}
\vspace{-1.5em}
%of our cross-domain strategy and the mainstream strategy
\begin{center}
\begin{tabular}{c | c : c | ccc | ccc}
\hline
\multirow{3}{*}{Budget} & \multicolumn{2}{c|}{\multirow{2}{*}{Proposed}} & \multicolumn{6}{c}{Performance(\%)}                    \\
\cline{4-9}
                        & \multicolumn{2}{c|}{}                            & \multicolumn{3}{c|}{Live} & \multicolumn{3}{c}{Cadaver} \\
\cline{2-9}
                        & External Background   & Background Inpainting   & mDSC   & mIOU   & mIoU\textsubscript{NB} & mDSC    & mIOU    & mIoU\textsubscript{NB}  \\
\hline
\hline
\multirow{4}{*}{1\%}    & \multicolumn{2}{c|}{No synthetic image}                        & 60.18  & 52.06  & 46.65  & 54.85   & 45.34   & 42.12   \\
                        & $\times$                    & $\checkmark$                     & 63.94  & 56.63  & 51.33  & 61.89   & 53.65   & 49.27   \\
                        & $\checkmark$                   & $\times$                       & 66.36  & \textbf{59.52}  & \textbf{53.26}  & \textbf{65.91}   & \textbf{58.51}   & 50.82   \\
                        & $\checkmark$                   & $\checkmark$                     & \textbf{66.71}  & 59.33  & 53.14  & 65.32   & 57.62   & \textbf{51.01}   \\
\hline
\multirow{4}{*}{10\%}   & \multicolumn{2}{c|}{No synthetic image}                          & 70.89  & 63.18  & 54.85  & 73.67   & 66.04   & 56.52   \\
                        & $\times$                     & $\checkmark$                     & 75.40  & 68.90  & \textbf{62.76}  & 76.34   & 69.71   & 60.90   \\
                        & $\checkmark$                   & $\times$                       & 75.53  & \textbf{70.47}  & 61.65  & 76.59   & 69.93   & 59.22   \\
                        & $\checkmark$                   & $\checkmark$                     & \textbf{75.91}  & 69.37  & 61.34  & \textbf{78.42}   & \textbf{72.05}   & \textbf{62.49}   \\
\hline
\multirow{4}{*}{50\%}   & \multicolumn{2}{c|}{No synthetic image}                          & 78.30  & 71.73  & 64.28  & 77.97   & 70.97   & 61.17   \\
                        & $\times$                     & $\checkmark$                     & 79.81  & 74.15  & 67.79  & 78.31   & 72.40   & 63.57   \\
                        & $\checkmark$                   & $\times$                       & 80.66  & 74.87  & 66.30  & 78.64   & 72.49   & 62.49   \\
                        & $\checkmark$                   & $\checkmark$                     & \textbf{80.99}  & \textbf{75.21}  & \textbf{67.91}  & \textbf{80.48}   & \textbf{74.43}   & \textbf{64.59}   \\
\hline
\multirow{4}{*}{100\%}  & \multicolumn{2}{c|}{No synthetic image}                         & 81.35  & 75.14  & 66.34  & 79.42   & 72.50   & 62.61   \\
                        & $\times$                     & $\checkmark$                     & 83.12  & 77.48  & 70.26  & 79.48   & 73.37   & \textbf{64.45}   \\
                        & $\checkmark$                   & $\times$                       & \textbf{83.64}  & \textbf{78.18}  & \textbf{70.86}  & 80.28   & 73.83   & 62.84   \\
                        & $\checkmark$                   & $\checkmark$                     & 82.72  & 77.18  & 69.11  & \textbf{80.51}   & \textbf{74.46}   & 63.46  \\
\hline
\multicolumn{9}{l}{The bold font indicates the best performance in each budget.}
\end{tabular}
\end{center}
\vspace{-2.5em}
\label{Tab_ext_bkgd}
\end{table*}

\end{document}